\documentclass{article}

\usepackage[preprint]{corl_2026}

\usepackage{pifont}

\usepackage{graphicx}
\usepackage{booktabs}

\usepackage{tikz} 
\usepackage{pgfplots} 
\usepgfplotslibrary{groupplots} 
\usetikzlibrary{calc} 
\definecolor{cleancolor}{RGB}{0,114,178} 
\definecolor{partialcolor}{RGB}{255,127,14}
\usepackage{graphicx} 
\usepackage{xcolor} 
\usepgfplotslibrary{groupplots}
\usetikzlibrary{patterns} 
\usepackage{amsmath} \usepackage{amssymb} 
\usetikzlibrary{patterns} 
\pgfplotsset{compat=1.18}
\usetikzlibrary{calc,patterns,plotmarks}
\usepackage{pgfplotstable}
\title{DAM-VLA: Decoupled Asynchronous Multimodal Vision Language Action model}

\usepackage{enumitem}

\usepackage[acronym,nomain]{glossaries}

\newacronym{vla}{VLA}{Vision-Language-Action}
\newacronym{vlm}{VLM}{Vision-Language Model}
\newacronym{gca}{GCA}{Gated Cross-Attention}
\newacronym{gru}{GRU}{Gated Recurrent Unit}
\newacronym{moe}{MoE}{Mixture of Experts}
\newacronym{dit}{DiT}{Diffusion Transformer}
\newacronym{ema}{EMA}{Exponential Moving Average}
\newacronym{adaln}{AdaLN}{Adaptive Layer Normalization}

\newacronym{ft}{F/T}{Force/Torque}
\newacronym{rgb}{RGB}{Red Green Blue}
\newacronym{dof}{DoF}{Degrees of Freedom}

\newacronym{laya}{LAYA}{Learning Asynchronous Multimodal Actions}
\newacronym{davla}{DAM-VLA}{Decoupled Asynchronous Multimodal Vision Language Action}
\newacronym{lfp}{LfP}{Learning from Play}
\newacronym{vam}{VAM}{Video-Action Model}

\newacronym{rq}{RQ}{Research Question}

\author{
  \textbf{Pankhuri Vanjani\textsuperscript{1}, Zhuoyue Li\textsuperscript{1*}, Jakub Suliga\textsuperscript{1*}, Moritz Reuss\textsuperscript{2},} \\
  \textbf{Gianluca Geraci\textsuperscript{1}, Xinkai Jiang\textsuperscript{1}, Rudolf Lioutikov\textsuperscript{1,3}} \\[4pt]
  \textsuperscript{1}Intuitive Robots Lab, Karlsruhe Institute of Technology (KIT), Germany \\
  \textsuperscript{2}NVIDIA \qquad \textsuperscript{3}Robotics Institute of Germany \\[2pt]
}

\begin{document}

\maketitle


\begin{abstract}
Vision-language-action (VLA) models inherit a shared synchronous clock from vision-language pretraining, processing every input at one rate. This is misaligned with physical interaction, where a high-frequency modality changes at hundreds of hertz, vision evolves more slowly, and language stays constant across an episode. A synchronous VLA oversamples slow modalities, undersamples fast ones, and caps action generation at the lowest effective frequency. We hypothesize that decoupling temporal processing per modality, letting each update and retain information at its own sensor rate, yields stronger representations and more robust control. We present \gls{davla}, which maintains per-modality latent buffers refreshed at sensor rates and read continuously by the action head, integrating new high-frequency modalities through gated cross-attention that leaves the pretrained backbone intact. Across seven contact-rich real-world manipulation tasks, \gls{davla} more than doubles the average success rate of the strongest synchronous baseline (95.2\% vs.\ 40.95\%) while sustaining smooth, reactive 100\,Hz control. Project website: \href{https://intuitive-robots.github.io/DAM-VLA/}{intuitive-robots.github.io/DAM-VLA/}

\end{abstract}



\section{Introduction}

Robot manipulation requires models that can act reliably across diverse tasks and environments. \gls{vla} models have emerged as a promising foundation for this goal, leveraging pretrained vision-language representations to generalize across tasks. They are building on imitation learning where representation quality drives robust behavior ~\cite{brohan2022rt,chi2025diffusion, vanjani2025disdp}. However, the architectural assumptions underlying these models were shaped by language and vision pretraining. In those domains a single, uniform processing clock is a natural fit.

Current \gls{vla} architectures are all \emph{synchronous} with respect to sensor modalities. At each fixed timestep, every input, vision, force-torque, proprioception, tactile, is encoded together and passed to the policy. This design is inherited from \glspl{vlm} optimized for language-grounded perception, not for real-time sensorimotor control. The result is a fundamental mismatch between the model's internal rhythm and the physical structure of the real world and its sensors.

This mismatch arises because heterogeneous sensors produce meaningful information at fundamentally different rates. As one concrete example, force-torque signals capture contact transients at 100-500 Hz, while RGB cameras provide informative signals at 3-10 Hz. Each modality also carries meaning over a different temporal horizon. A force spike is relevant for milliseconds, while a scene layout remains stable across seconds. A synchronous fixed-clock model is doubly misaligned: it ignores both the natural sensor rate at which each modality becomes informative and the natural horizon over which its information remains relevant.
This mismatch produces three major challenges.
First, \textbf{Redundant compute}: the expensive \gls{vlm} encoder re-processes semantically identical frames every step, wasting the compute budget needed for higher action rates.
Second, \textbf{Cross-modal rate mismatch}: a uniform clock simultaneously undersamples fast-updating modalities and oversamples slow-updating ones. No modality is processed at the rate its signal structure demands. Third, \textbf{Action latency}: policy execution is gated on the arrival of a fully synchronized observation bundle. Waiting for the slowest modality directly increases latency and reduces effective control frequency.

\begin{figure}[t]
    \centering
    \includegraphics[width=\textwidth]{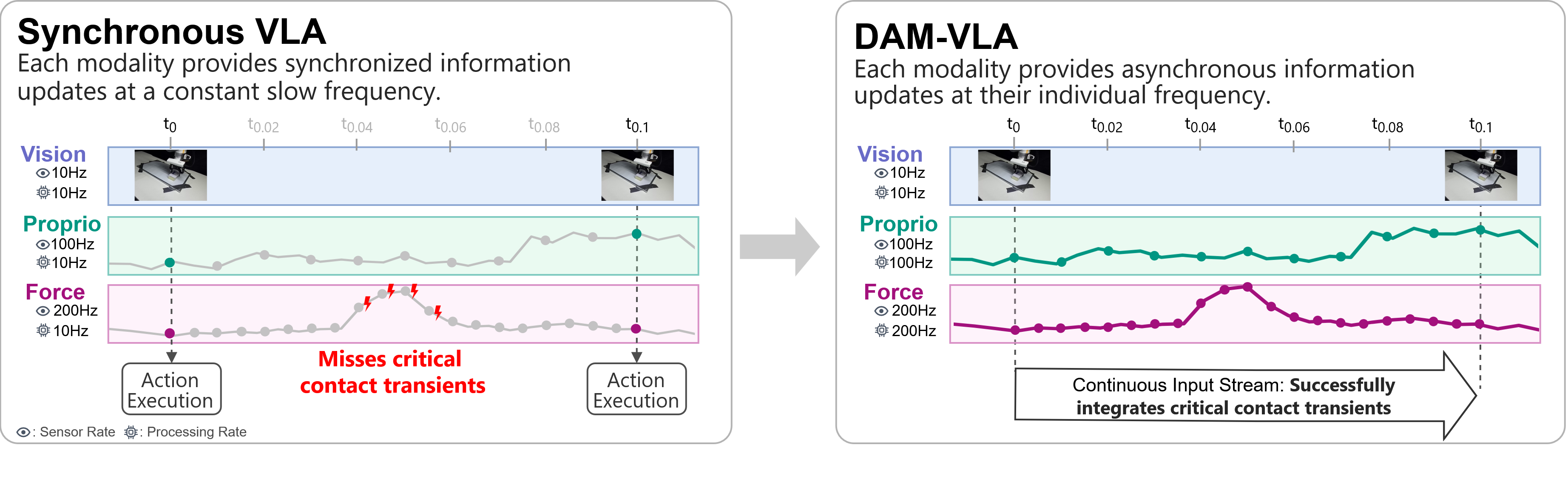}
    \caption{
    Standard synchronous VLAs operate on a single slow clock, missing critical high-frequency contact transients. In contrast, \gls{davla} updates each modality asynchronously at its natural sensor rate, successfully capturing fast dynamics and enabling smooth, continuous control.
    }

    \label{fig:concept_figure_clean}
\end{figure}

\emph{Decoupling} is the right architectural principle for multimodal robot learning. Rather than forcing all modalities into a synchronous token stream, each modality updates, retains, and contributes information according to its own temporal dynamics. This principle spans two dimensions. 1) \textbf{asynchrony}: each modality is processed at its natural sensor instead of down- or upsampling it to a shared clock. 2) \textbf{temporal context}: each modality maintains a buffer sized to its meaningful horizon. Short- and long-lived signals are preserved at their appropriate timescale. These dimensions lead to representations that reflect the information structure of the sensors instead of synchronous projections.

Following these insights we present \gls{davla}, a decoupled asynchronous \gls{vla} architecture for dexterous robot manipulation. In this work the heterogeneous modality set comprises visual streams and forces/torques as a representative high-frequency modality updated at the natural sensor rate. The architecture is not specific to this sensor set. Other modalities can be leveraged at their sensor rates via independent memory buffers.
\textbf{Our contributions are:}
\begin{enumerate}[leftmargin=*, itemsep=0pt, topsep=0pt, before=\vspace{-\parskip},after=\vspace{-\parskip}]
    \item \textbf{Asynchronous multimodal architecture}: We introduce a decoupled processing design in which each modality stream updates independently at its natural frequency, with per-modality temporal context windows sized to the meaningful horizon of that signal.
    \item \textbf{Improved performance through asynchronous representations}: By preserving the natural information structure of each sensor, \gls{davla} learns multimodal representations that lead to higher task success rates than synchronous baselines.
    \item \textbf{Reduced inference latency}: Decoupling action generation from slow modality update cycles, such as periodic \gls{vlm} re-encoding, enables the policy to act continuously at the control frequency, reducing end-to-end latency and increasing effective control frequency.
\end{enumerate}

We evaluate \gls{davla} on seven contact-rich, real world manipulation tasks, using X-VLA \cite{zheng2025x} as the \gls{vla} backbone, with force/torque as the high-frequency modality. The results show that asynchronous decoupled processing improves average success rate by \textbf{54.25\%} while running smoothly at $100\mathrm{Hz}$ demonstrating, the practical value of multi-rate \gls{vla} design.

\section{Related Work}

\textbf{Asynchronous and Efficient VLA Inference.}
Early work on asynchronous robot learning addresses the mismatch between slow \gls{vla} inference and real-time control demands. Black et al.~\cite{black2026real} and \gls{vla}-RAIL~\cite{zhao2025vla} overlap chunk generation and execution to maintain continuity, while A2C2~\cite{sendai2025leave} attaches a lightweight correction head that injects time-aware residuals at every control step, recovering reactivity under inference delay.  A complementary slow-fast thread splits the model into a semantic \gls{vlm} pathway and a fast action pathway: FiS-\gls{vla}~\cite{chen2026fast} shares parameters between both at a fixed 1:4 frequency ratio. DuoCore~\cite{zou2025asynchronous} bridges them via a latent buffer achieving 30Hz whole-body control. On the efficiency side, \gls{vla}-Cache~\cite{xu2025vla} and SD-VLA~\cite{qiu2026efficient} avoid recomputing static visual tokens across frames ,Realtime-\gls{vla}~V2~\cite{yang2026realtime} focuses on system level optimizations, and VLASH~\cite{tang2025vlash},  FASTER~\cite{lu2026faster} address temporal misalignment between prediction and execution.

These works either treat asynchrony as a system-level scheduling problem or focus on a single fast-slow split at a fixed ratio. All assume a synchronized observation bundle is available at each inference cycle. We study a complementary problem: heterogeneous sensory streams that update at different rates. None consider the general case of heterogeneous sensors streaming at different rates, nor do they study how the choice of modality integration mechanism affects the quality of the learned multimodal representation. Rather than requiring complete observations at every step, \gls{davla} maintains per-modality latent buffers updated at sensor rates, addressing both issues jointly.
Outside the \gls{vla} setting, ManipForce~\cite{lee2025manipforce} shows that training with native async RGB and force/torque streams outperforms downsampled baselines, directly motivating our multi-rate approach.

\textbf{Multimodal Sensing and Temporal Context in \glspl{vla}.}
Recent work has extended \gls{vla} architectures with additional sensing modalities for contact-rich manipulation, with force and tactile sensing as the most commonly integrated examples. TA-\gls{vla}~\cite{zhang2025ta} and ForceVLA2~\cite{li2026forcevla2} inject torque at different architectural points, the latter adding force-based prompts into the \gls{vlm} for hybrid force-position control. TacVLA~\cite{zhang2026tacvla} fuses tactile tokens with a hard contact gate, while FAVLA~\cite{li2026favla} uses the slow \gls{vlm} to predict near-future force variation to schedule the fast action expert. FD-\gls{vla}~\cite{zhao2026fd} takes the opposite extreme, distilling force entirely from vision without a physical sensor. Across all these works, force signals are fused synchronously into a shared token stream. 
To handle temporally extended tasks, recent \glspl{vla}~\cite{torne2026mem, sridhar2025memer, zheng2025tracevla,lin2025hif, koo2025hamlet, shi2025memoryvla, lin2025echovla, ni2025swiftvla, dai2026robomme, gao2026gated} integrate latent memory through recurrent encoders, memory banks, or attention-based compressors, yet universally enforce a synchronous update regime that ties all sensor streams to a single clock. 
\gls{davla} addresses both. Each modality is maintained as an independently buffered asynchronous stream updated at its sensor rate, and integrated via \gls{gca} pathways with gating strategies matched to each modality's signal structure.
We use force/torque as a new, high-frequency modality.

\section{Method}

\subsection{Problem Formulation}

We consider a policy that receives observations from $M$ 
heterogeneous sensing modalities $\mathcal{M} = \{m_1, \ldots, m_M\}$, each 
producing meaningful information at distinct rates. In our experiments, $\mathcal{M}$ contains a third person scene camera at $25\mathrm{Hz}$, a wrist mounted camera at $25\mathrm{Hz}$, force/torque at $100\mathrm{Hz}$ sourced directly from the Franka's internal sensor, proprioception at $100\mathrm{Hz}$ and a language  instruction that is static for the duration of each episode.
A synchronous \gls{vla} queries all modalities at a single control frequency, treating every observation as if it contains every relevant information since the last step.
This is suboptimal for heterogeneous sensing. It oversamples redundant visual data, undersamples fast contact transients, and blocks action generation until a complete observation bundle is available.
We decouple each modality's update rate from the control rate, maintaining a per modality latent context $z_{\mathrm{m}}$ in a shared buffer.
Between update events, the cached context is held as a latent buffer and read by the action head at every control step. 
The action head is conditioned on this latent buffer, so that action generation is not blocked by individual modality's update rate.

\begin{figure}[t]
    \centering
    \includegraphics[width=\textwidth]{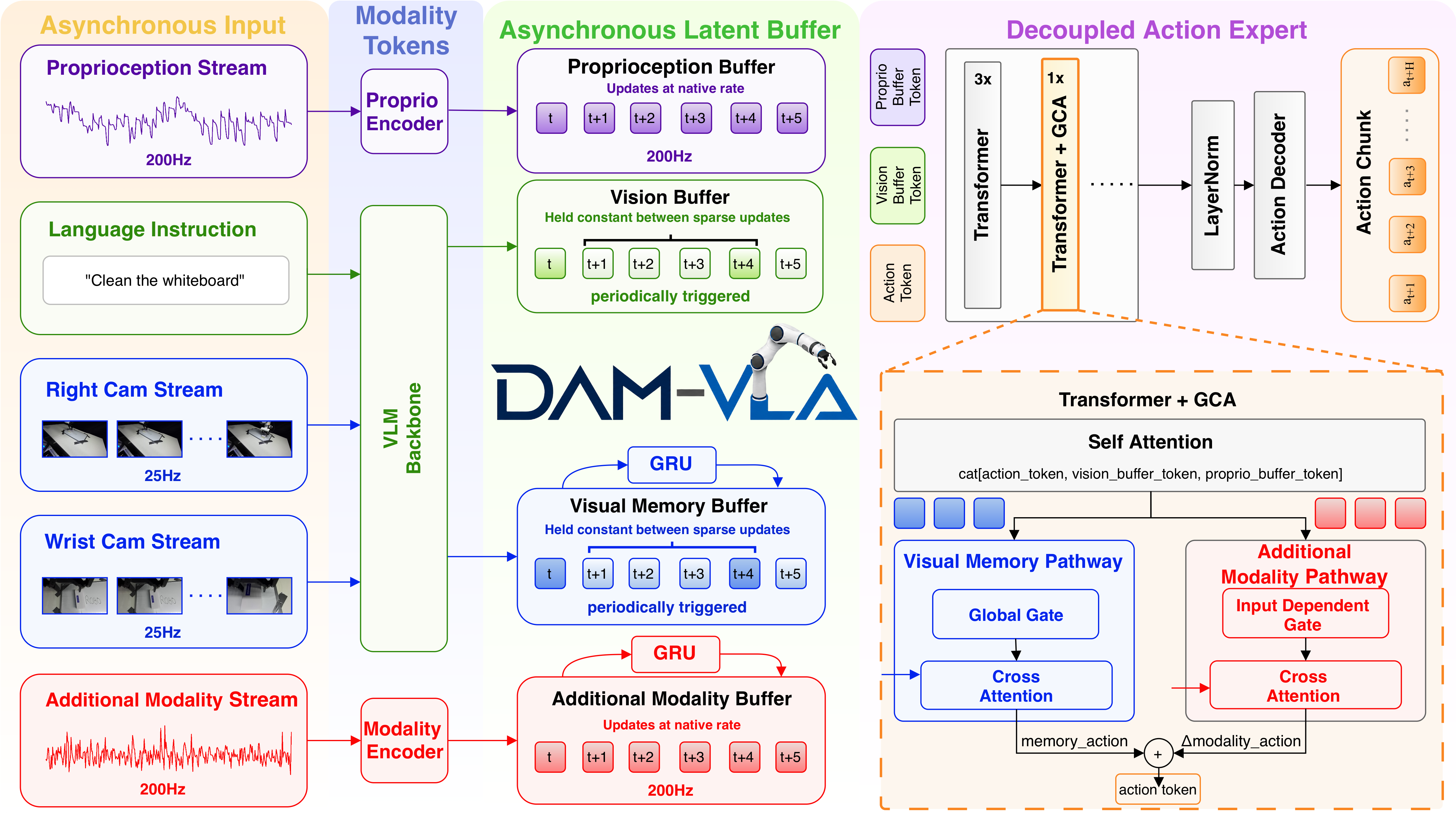}
\caption{\textbf{\gls{davla} architecture.} Each modality stream encodes tokens into independent latent buffers at their sensor rate: vision periodically, proprioception and force/torque at high frequency. The action expert reads all buffers continuously via parallel \gls{gca} pathways, 
a global-gate pathway for visual memory and an input-dependent gate pathway for force/torque, adding new modalities through dedicated cross-attention modules that preserve the pretrained self-attention structure.}
    \label{fig:Architecture}
\end{figure}

\subsection{\gls{davla} Architecture}

\paragraph{Asynchronous Data Collection.}

Rather than synchronising all sensor streams to a single timestamp before recording, we collect each modality independently at its sensor rate and store observations with per-modality timestamps. During training, we construct the observation context for each action label by fetching a fixed history window from each modality at its natural temporal resolution: sparse visual history of 16 frames at $25\textrm{Hz}$, corresponding to approximately 0.64 s of semantic visual context and dense proprioceptive and force/torque histories of 96 samples each at $100\textrm{Hz}$, corresponding to approximately 0.96 s of high-resolution state and contact history which captures fine-grained short-horizon dynamics. Vision provides semantic context over a longer horizon while force and proprioception provide high-resolution contact and state information over shorter periods.

\paragraph{Multimodal asynchronous Latent buffer}
Continuous re-encoding of all modalities at full control frequency is wasteful. 
Vision changes slowly between manipulation phases, while force at contact change an order of magnitude faster. 
We maintain a shared latent buffer $\mathcal{B}\!=\!\{Z^m\}_{m \in \mathcal{M}}$ that holds one token sequence per modality. Each $Z^m \in \mathbb{R}^{N_m \times d}$ is refreshed at a modality-specific rate.
At every inference step the tokens of each $Z^m$ are processed but not consumed, i.e., if the inference frequency is higher than the sensor update the same sensor input is read again.
The action head reads the entire buffer at every inference step, fully decoupling action generation from any individual modality's encoding schedule.  This work uses these modalities:
\\\textit{1) Language tokens}: For each episode a language instruction is encoded once at the beginning.
\\\textit{2) Visual tokens}: The primary camera and the wrist camera are encoded by a vision-language encoder to generate a sequence of patch tokens. To avoid redundant re-encoding of semantically identical frames, the visual encoder is invoked every 4 inference steps to get \gls{vlm} embeddings.
\\\textit{3) Proprioception tokens}: Joint states are concatenated with the action tokens as in X-VLA~\cite{zheng2025x}, but read at the full 100$\mathrm{Hz}$ control rate, then encoded and processed in the asynchronous latent buffer.
\\\textit{4) Force tokens}: Joint-torque readings arrive at the full control frequency and are smoothed via exponential moving average before being accumulated in a rolling buffer. 
A \gls{gru} encodes this buffer and cross-attention over force registers compresses it to $Z^{ft}$, updated every control step independently of the visual schedule.

\gls{davla} also maintains a short-term visual memory for context across sparse updates: each visual update appends the new frame embedding to a rolling buffer of the $K$ most recent ones, which a \gls{gru} encodes and learned-query cross-attention compresses to $N_{\mathrm{mem}}$ tokens, producing $Z^{\mathrm{mem}}$. Summarizing $K$ frames rather than one snapshot, $Z^{\mathrm{mem}}$ stays valid while held constant between multiple inference updates.

\begin{table}[b!]
\centering
\small
\begin{tabular}{llcccc}
\toprule
\textbf{Configuration} & \textbf{Isolates} & \textbf{Async} & \textbf{Force} & \textbf{Mem.} & \textbf{Integ.} \\
\midrule

\cmidrule(lr){1-6}
X-VLA$_{25}$  & std.\ VLA regime (25\,Hz) & \ding{55} & \ding{55} & \ding{55} & --  \\
X-VLA$_{100}$ & naive high-freq.\ (100\,Hz) & \ding{55} & \ding{55} & \ding{55} & --  \\
\cmidrule(lr){1-6}
\cmidrule(lr){1-6}
X-VLA$_{A\!F\!M}$        & concat.\ baseline       & \ding{51} & \ding{51} & \ding{51} & concatenate       \\
\gls{davla}$_{/\!F\!/\!M}$ & async.\ alone           & \ding{51} & \ding{55} & \ding{55} & --        \\
\gls{davla}$_{/\!F}$   & memory contribution     & \ding{51} & \ding{55} & \ding{51} & \gls{gca} \\
\gls{davla}$_{/\!M}$   & force contribution      & \ding{51} & \ding{51} & \ding{55} & \gls{gca} \\
\gls{davla} (Ours)   & full model              & \ding{51} & \ding{51} & \ding{51} & \gls{gca} \\
\bottomrule
\end{tabular}
\caption{Evaluated configurations. All share the X-VLA backbone~\cite{zheng2025x}, training data, and task split. \textit{cat}: modality tokens concatenated into the input sequence; \textit{\gls{gca}}: our gated cross-attention.}
\label{tab:configs}
\end{table}

\paragraph{Dual-pathway modulation via gated cross-attention.}
Inputs from all modalities are jointly processed in the action expert of the \gls{vla}. 
Most naive baselines concatenate these heterogeneous information streams into a single flat sequence. This design, inherited from \gls{vlm} token processing, is ill-suited to asynchronous updates and information decoupling. Also it forces new modality tokens into pretrained self-attention weights that have never seen them, risking corruption of pretrained representations.
Instead, \gls{davla} augments the standard token-sequence input with two novel conditioning mechanisms: parallel \gls{gca} pathways that modulate the action expert on memory and force, inserted at every four transformer block of the action expert.
This preserves pretrained weights untouched while allowing new modalities to inject residual corrections onto action tokens only. This is inspired by \cite{alayrac2022flamingo}.

\textit{Visual Memory pathway.} Compressed visual memory tokens $Z^{\mathrm{mem}}$
condition action tokens $Z^{(\ell)}$ via a zero-initialized residual:
\begin{equation}
  Z^{(\ell+1)} = Z^{(\ell)}
    + \tanh(\alpha)\;\mathrm{CA}\!\bigl(\mathrm{LN}(Z^{(\ell)}),\;
      Z^{\mathrm{mem}}\bigr),
\end{equation}
where $\alpha$ is a learned scalar initialized to zero so the pathway
contributes nothing at the start of training and grows gradually without
disrupting pretrained representations. 
A global gate is appropriate here
because temporal visual context is relevant throughout the complete episode.

\textit{Additional modality pathway.}: Other modalities use the same insertion points but an input-dependent gate, which we demonstrate with force tokens $Z^{\mathrm{ft}}$.
\begin{equation}
  Z^{(\ell+1)} = Z^{(\ell)}
    + \sigma\!\bigl(W\,\bar{z}^{\mathrm{ft}}\bigr)\;
      \mathrm{CA}\!\bigl(\mathrm{LN}(Z^{(\ell)}),\;
      Z^{\mathrm{ft}}\bigr),
\end{equation}
where $\bar{z}^{\mathrm{ft}}$ is the mean-pooled force token and the
sigmoid gate is initialized near-closed. 
A static gate would be driven open by contact-phase gradients and closed by free-space gradients, converging to a compromise that either leaks noise during free motion or under-weights force during contact. 
The input-dependent gate lets the network learn \emph{when} force is informative without requiring explicit contact detection.
Critically, force queries the pre-memory-update action tokens $Z^{(\ell)}$  rather than the memory-updated tokens, computing a pure additive delta
\begin{equation}
  \Delta^{\mathrm{ft}} = \mathrm{CA}\!\bigl(\mathrm{LN}(Z^{(\ell)}),\;
  Z^{\mathrm{ft}}\bigr) - Z^{(\ell)},
\end{equation}
that is added on top of the memory update, keeping the two conditioning pathways orthogonal and preventing cross-modal entanglement. The force gate responds to raw contact state, not to a signal already mixed with visual memory context to ensure reactive response at high frequency.

\paragraph{Training- and inference-time asynchrony}
During training, all modalities are first aligned to a common 100\,Hz timeline for consistent action labeling. Visual observations are then sampled with stride $S$, recovering a sparse history matching the camera's sensor rate, while force is sampled consecutively at the full 100\,Hz. This mirrors the inference-time buffer, where vision updates sparsely and force updates every control step.
We use asynchronous, delay-aware execution~\cite{shukor2025smolvla} and characterize horizon-dependent replanning, as studied for X-VLA in~\cite{lu2026faster}.

\begin{table}[t!]
\centering
\resizebox{\columnwidth}{!}{
\begin{tabular}{lccccccc|c}
\toprule
\textbf{Model} 
& \textbf{Scarf} 
& \textbf{Whiteboard} 
& \textbf{Button} 
& \textbf{Handwash} 
& \textbf{Lego} 
& \textbf{Socket} 
& \textbf{Sweep} 
& \textbf{Avg.} \\
\midrule
\multicolumn{9}{l}{\textit{Baselines}} \\
\hline
X-VLA$_{25}$                & 80.0  & 86.7  & 13.3 & 0.0  & 0.0  & 6.7  & 100.0 & 40.95 \\
X-VLA$_{100}$               & 80.0  & 13.3  & 6.7  & 0.0  & 0.0  & 0.0  & 53.3  & 21.9 \\
\hline
\multicolumn{9}{l}{\textit{Ours}} \\
\hline
X-VLA$_{AFM}$               & 100.0 & 73.3  & 13.3 & 86.7 & 0.0  & 6.7  & 100.0 & 54.3 \\
\gls{davla}$_{/\!F/M}$      & 80.0  & 66.7  & 40.0 & 20.0 & 0.0  & 6.7  & 66.7  & 40.0 \\
\gls{davla}$_{/F\!}$        & 100.0 & 73.3  & 86.7 & 40.0 & 0.0  & 6.7  & 100.0 & 58.1 \\
\gls{davla}$_{/M\!}$        & 100.0 & 86.7  & 86.7 & 80.0 & 13.3 & 13.3 & 86.7  & 66.7 \\
\textbf{\gls{davla} (Ours)} & \textbf{100.0} & \textbf{100.0} & \textbf{93.3} & \textbf{100.0} & \textbf{93.3} & \textbf{80.0} & \textbf{100.0} & \textbf{95.2} \\
\bottomrule
\end{tabular}}
\caption{Task success rate (\%) per configuration over 15 trials each. Best result per column in \textbf{bold}.}
\label{tab:main}
\end{table}

\section{Evaluation}
\label{sec:result}

We evaluate \gls{davla}, our proposed asynchronous decoupled Multimodal VLA with per-modality latent buffering, against synchronous baselines across a suite of manipulation tasks.
Our evaluation is structured around four research questions: \textbf{(RQ1)} Does synchronous processing limit manipulation performance, and does naive frequency scaling resolve this? 
\textbf{(RQ2)} Does asynchronous decoupling affect task performance?
\textbf{(RQ3)} Can high frequency modalities, e.g. forces, combined with individual memory improve performance?
\textbf{(RQ4)} Does the integration mechanism for additional modalities affect task performance in a decoupled asynchronous backbone?

\subsection{Experimental Setup}
We conduct experiments on a Franka Emika Panda arm using a fixed third-person scene camera and a wrist-mounted camera. RGB streams are recorded at 25 Hz and proprioception at 100Hz. Force/torque readings are sourced directly from the Franka's internal sensor at 100 Hz. 
We evaluate on seven manipulation tasks spanning a range of contact requirements,
reactivity demands, and precision constraints:
\textit{Scarf} folding, \textit{Whiteboard} cleaning, \textit{Button} pressing, \textit{Handwash} top press, \textit{Socket} insertion, \textit{Sweep} beads into a dustpan, and \textit{Lego} piece arranging.

We measure \textbf{(i)} task success rate (\%) over 15 trials per task and  
\textbf{(ii)} average episode length (seconds), compared against teleoperation demonstrations as a natural-pacing reference. In the Appendix we further compare and show improvements in trajectory smoothness across two smoothness metrics.

\paragraph{Baselines and Ablations}

We evaluate six configurations (Table~\ref{tab:configs}), each isolating one design axis of \gls{davla}, all built on the same X-VLA backbone~\cite{zheng2025x} and finetuned with identical data and splits. 
The X-VLA baselines test the standard synchronous approach and a naive high-frequency variant that upsamples visual frames to $100\mathrm{Hz}$ without redundancy suppression or temporal memory. 
X-VLA$_{AFM}$ is the strongest concatenation baseline. It accesses the same force and memory information as \gls{davla} but integrates it via a flat token sequence in contrast to \gls{davla}'s gated cross-attention. 
The remaining ablations isolate the individual contributions of asynchronous decoupling (\gls{davla}$_{/F/M}$), force (\gls{davla}$_{/F}$), and visual memory (\gls{davla}$_{/M}$).

All reported results run on a $100$~Hz controller: the X-VLA$_{25}$ and X-VLA$_{100}$ baselines replan at $\approx\!1$ and $\approx\!3.5$~Hz, and \gls{davla} ($s{=}22$) at $\approx\!5.5$~Hz with smooth, stable execution. To probe the frequency limit, we additionally ran \gls{davla} on a $200$~Hz controller, where it stayed reactive from $\approx\!8$~Hz ($s{=}22$) up to $\approx\!17$~Hz ($s{=}6$). Force and proprioception enter the buffer at $200$~Hz and are read every inference step, decoupling the input rate from replanning.

\subsection{Result Analysis}

\subsubsection*{ RQ1: Synchronous processing hits a performance ceiling despite frequency scaling.}
\label{sec:rq1}

\begin{figure}[t]
    \centering
    \includegraphics[width=\textwidth]{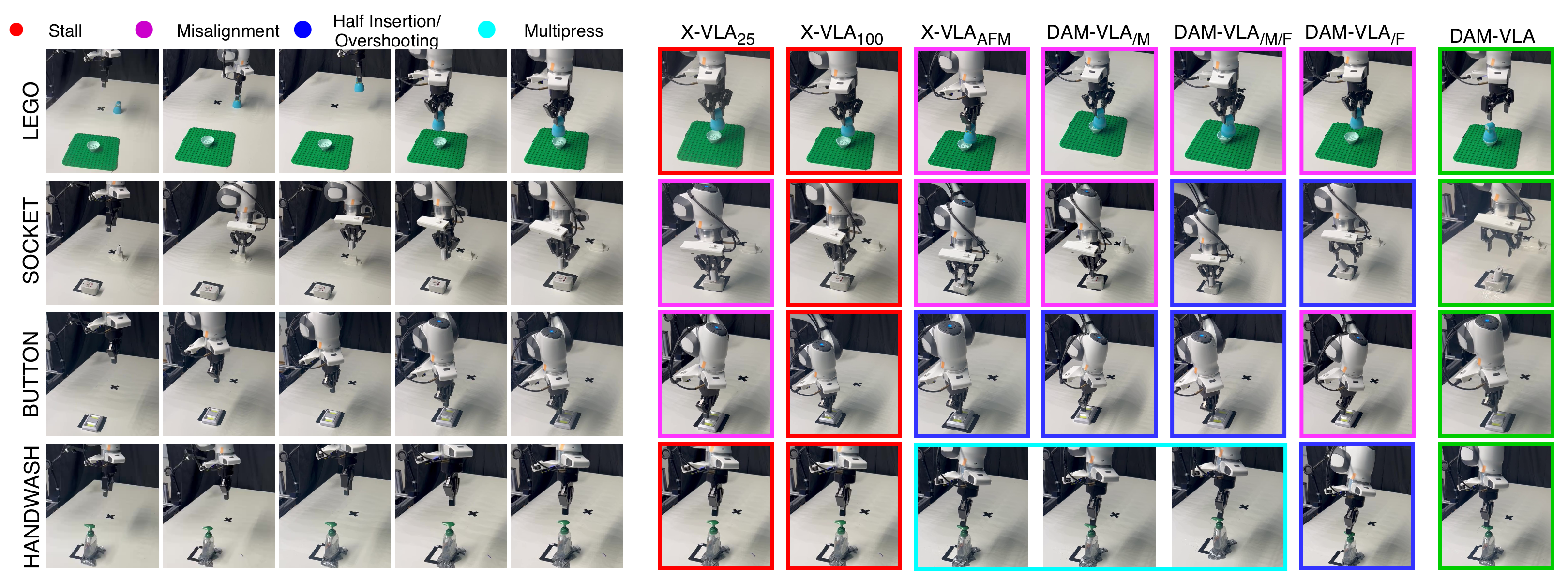}
    \caption{Qualitative rollout comparison across a subset of manipulation tasks. green boxes indicate successful \gls{davla} executions and the other colored boxes indicate $X$-VLA$_{25}$ partial outcomes or the failure modes.}
    \label{fig:qualitative results}
\end{figure}

Table~\ref{tab:main} shows that X-VLA\textsubscript{25} achieves an average success rate of 40.95\%, with strong performance on visually-guided tasks such as whiteboard cleaning (86.7\%) and sweep (100\%), but near-zero performance on tasks requiring precise contact: handwash (0\%), Lego (0\%), and socket insertion (6.7\%). Increasing control frequency to 100\,Hz drops average success further to 21.9\%, with degradation across almost all tasks. Notably, sweep falls from 100\% to 53.3\% and whiteboard 
from 86.7\% to 13.3\%, showing that higher frequency actively hurts even on tasks where the synchronous baseline was strong. X-VLA\textsubscript{100} was trained with visual observations naively upsampled to match the 100\,Hz control rate. At this rate, identical frames are paired with different action labels, creating a contradictory training signal that causes the policy to predict small hesitant movements rather than committing to actions. This redundant-frame bias stalls execution and produces jerky motion (Fig.\ref{fig:qualitative results}). These results confirm that naive frequency scaling is not a viable path to high-frequency manipulation, and that the bottleneck is architectural rather than a matter of controller rate.

\subsubsection*{RQ2: Asynchronous decoupling alone recovers performance lost by naive frequency scaling.}

\label{sec:rq2} 

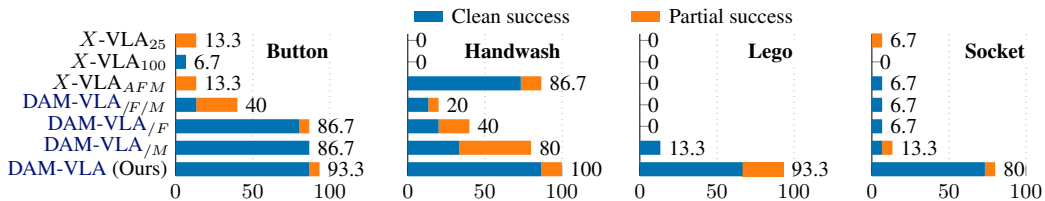
\begin{figure*}[b!]
    \centering
    \resizebox{\textwidth}{!}{%
        \begin{tikzpicture}

        \begin{groupplot}[
            group style={
                group size=4 by 1,
                horizontal sep=1.05cm,
            },
            width=4.0cm,
            height=3.7cm,
            xbar stacked,
            xmin=0,
            xmax=105,
            xtick={0,50,100},
            xmajorgrids=true,
            grid style={dotted, gray!70},
            axis x line*=bottom,
            x axis line style={draw=none},
            axis y line*=left,
            /pgf/bar width=6pt,
            tick label style={font=\small},
            ytick=data,
            y dir=reverse,
            symbolic y coords={
                xvla25,
                xvla100,
                xvlaafm,
                davlafm,
                davlaf,
                davlam,
                davlafull
            },
            enlarge y limits=0.05,
            clip=false,
        ]

        \nextgroupplot[
            yticklabels={
                $X$-VLA$_{25}$,
                $X$-VLA$_{100}$,
                $X$-VLA$_{AFM}$,
                \gls{davla}$_{/\!F/M}$,
                \gls{davla}$_{/F\!}$ ,
                \gls{davla}$_{/M\!}$,
                \gls{davla} (Ours)
            }
        ]

        \addplot+[fill=cleancolor, draw=none] coordinates {
            (0,xvla25)
            (6.67,xvla100)
            (0,xvlaafm)
            (13.33,davlafm)
            (80,davlaf)
            (86.67,davlam)
            (86.67,davlafull)
        };

        \addplot+[fill=partialcolor, draw=none] coordinates {
            (13.33,xvla25)
            (0,xvla100)
            (13.33,xvlaafm)
            (26.67,davlafm)
            (6.67,davlaf)
            (0,davlam)
            (6.67,davlafull)
        };

        \node[anchor=west, font=\small] at (axis cs:13.33,xvla25) {13.3};
        \node[anchor=west, font=\small] at (axis cs:6.67,xvla100) {6.7};
        \node[anchor=west, font=\small] at (axis cs:13.33,xvlaafm) {13.3};
        \node[anchor=west, font=\small] at (axis cs:40,davlafm) {40};
        \node[anchor=west, font=\small] at (axis cs:86.67,davlaf) {86.7};
        \node[anchor=west, font=\small] at (axis cs:86.67,davlam) {86.7};
        \node[anchor=west, font=\small] at (axis cs:93.33,davlafull) {93.3};

        \node[
            anchor=north east,
            font=\bfseries\small,
            fill=white,
            fill opacity=0.9,
            text opacity=1,
            inner sep=1.5pt
        ] at (rel axis cs:0.97,0.96) {Button};

        \nextgroupplot[
            yticklabels={,,,,,,}
        ]

        \addplot+[fill=cleancolor, draw=none] coordinates {
            (0,xvla25)
            (0,xvla100)
            (73.33,xvlaafm)
            (13.33,davlafm)
            (20,davlaf)
            (33.33,davlam)
            (86.67,davlafull)
        };

        \addplot+[fill=partialcolor, draw=none] coordinates {
            (0,xvla25)
            (0,xvla100)
            (13.33,xvlaafm)
            (6.67,davlafm)
            (20,davlaf)
            (46.67,davlam)
            (13.33,davlafull)
        };

        \node[anchor=west, font=\small] at (axis cs:0,xvla25) {0};
        \node[anchor=west, font=\small] at (axis cs:0,xvla100) {0};
        \node[anchor=west, font=\small] at (axis cs:86.67,xvlaafm) {86.7};
        \node[anchor=west, font=\small] at (axis cs:20,davlafm) {20};
        \node[anchor=west, font=\small] at (axis cs:40,davlaf) {40};
        \node[anchor=west, font=\small] at (axis cs:80,davlam) {80};
        \node[anchor=west, font=\small] at (axis cs:100,davlafull) {100};

        \node[
            anchor=north east,
            font=\bfseries\small,
            fill=white,
            fill opacity=0.9,
            text opacity=1,
            inner sep=1.5pt
        ] at (rel axis cs:0.97,0.96) {Handwash};

        \nextgroupplot[
            yticklabels={,,,,,,}
        ]

        \addplot+[fill=cleancolor, draw=none] coordinates {
            (0,xvla25)
            (0,xvla100)
            (0,xvlaafm)
            (0,davlafm)
            (0,davlaf)
            (13.33,davlam)
            (66.67,davlafull)
        };

        \addplot+[fill=partialcolor, draw=none] coordinates {
            (0,xvla25)
            (0,xvla100)
            (0,xvlaafm)
            (0,davlafm)
            (0,davlaf)
            (0,davlam)
            (26.67,davlafull)
        };

        \node[anchor=west, font=\small] at (axis cs:0,xvla25) {0};
        \node[anchor=west, font=\small] at (axis cs:0,xvla100) {0};
        \node[anchor=west, font=\small] at (axis cs:0,xvlaafm) {0};
        \node[anchor=west, font=\small] at (axis cs:0,davlafm) {0};
        \node[anchor=west, font=\small] at (axis cs:0,davlaf) {0};
        \node[anchor=west, font=\small] at (axis cs:13.33,davlam) {13.3};
        \node[anchor=west, font=\small] at (axis cs:93.33,davlafull) {93.3};

        \node[
            anchor=north east,
            font=\bfseries\small,
            fill=white,
            fill opacity=0.9,
            text opacity=1,
            inner sep=1.5pt
        ] at (rel axis cs:0.97,0.96) {Lego};

        \nextgroupplot[
            yticklabels={,,,,,,}
        ]

        \addplot+[fill=cleancolor, draw=none] coordinates {
            (0,xvla25)
            (0,xvla100)
            (6.67,xvlaafm)
            (6.67,davlafm)
            (6.67,davlaf)
            (6.67,davlam)
            (73.33,davlafull)
        };

        \addplot+[fill=partialcolor, draw=none] coordinates {
            (6.67,xvla25)
            (0,xvla100)
            (0,xvlaafm)
            (0,davlafm)
            (0,davlaf)
            (6.67,davlam)
            (6.67,davlafull)
        };

        \node[anchor=west, font=\small] at (axis cs:6.67,xvla25) {6.7};
        \node[anchor=west, font=\small] at (axis cs:0,xvla100) {0};
        \node[anchor=west, font=\small] at (axis cs:6.67,xvlaafm) {6.7};
        \node[anchor=west, font=\small] at (axis cs:6.67,davlafm) {6.7};
        \node[anchor=west, font=\small] at (axis cs:6.67,davlaf) {6.7};
        \node[anchor=west, font=\small] at (axis cs:13.33,davlam) {13.3};
        \node[anchor=west, font=\small] at (axis cs:80,davlafull) {80};

        \node[
            anchor=north east,
            font=\bfseries\small,
            fill=white,
            fill opacity=0.9,
            text opacity=1,
            inner sep=1.5pt
        ] at (rel axis cs:0.97,0.96) {Socket};

        \end{groupplot}

\node[
    anchor=south,
    fill=white,
    inner sep=1pt
] at ($(group c1r1.north west)!0.5!(group c4r1.north east)+(0,0.08cm)$) {
    \begin{tikzpicture}
        \draw[fill=cleancolor, draw=none] (0,0) rectangle (0.42,0.16);
        \node[anchor=west, font=\footnotesize] at (0.52,0.08) {Clean success};

        \draw[fill=partialcolor, draw=none] (3.25,0) rectangle (3.67,0.16);
        \node[anchor=west, font=\footnotesize] at (3.77,0.08) {Partial success};
    \end{tikzpicture}
};

        \end{tikzpicture}
    }
    \caption{Success rates across different tasks and model configurations. Blue indicates clean success, while orange indicates partial success.}
    \label{fig:partial}
\end{figure*}

\gls{davla}$_{/F/M}$ isolates the effect of asynchronous 
decoupling without any additional modality. It runs at 
100\,Hz with vision updated sparsely and proprioception 
updated at every control step. It achieves 40.0\% average 
success, nearly matching X-VLA\textsubscript{25} at 40.95\% 
while running at higher frequency, and outperforming 
X-VLA\textsubscript{100} at 21.9\%. At inference, vision is 
cached than re-encoded every step, avoiding the 
execution collapse seen in X-VLA\textsubscript{100}. 
Decoupling also begins to unlock tasks that synchronous 
baselines could not do at all: button press improves from 
13.3\% to 40.0\% and handwash from 0.0\% to 20.0\%. Some regression remains on visually-guided tasks like sweep (66.7\% vs 100.0\%), since no memory pathway preserves visual context between sparse updates. 

\subsubsection*{RQ3: Additional modalities at their sensor rates further improve the decoupled foundation.}
\label{sec:rq3}

Building on the decoupled foundation of \gls{davla}$_{/F/M}$ (40.0\%), every configuration that adds modality information improves: X-VLA$_{AFM}$ (54.3\%), \gls{davla}$_{/F}$ (58.1\%), \gls{davla}$_{/M}$ (66.7\%), and the full \gls{davla} (95.2\%)
This confirms the decoupled architecture is a foundation additional modalities can build on, each contributing independently.
Memory alone provides meaningful gains over the decoupled baseline. \gls{davla}$_{/F}$ achieves 100\% on scarf and sweep and 86.7\% on button press, but falls short on contact-critical tasks: partial presses without depth regulation on button, over-pressing that spills liquid on handwash (Fig.~\ref{fig:partial}), and 0\% on Lego despite reaching the target (Fig.~\ref{fig:qualitative results}). Force alone also improves over the decoupled baseline but introduces distinct failures without memory. \gls{davla}$_{/M}$ achieves 86.7\% on button and 80.0\% on handwash, but in 46.67\% of rollouts presses repeatedly, unable to retain that contact was made, and on Lego overshoots, failing to align.
The full model resolves both failure modes by combining force-guided contact with memory-stabilized sequencing.
On sweep, \gls{davla}$_{/F}$ achieves 100\% but takes 
40\,s per episode versus 22.5\,s for \gls{davla}, slowing 
at every contact boundary without force. Memory stabilizes 
sequential context and prevents repetitions. Force 
provides the contact signal needed to terminate interactions 
precisely. Together they reach 93.3\% on Lego and 80.0\% on 
socket where all synchronous baselines scored near zero.

\begin{figure*}[t]
    \centering
    \includegraphics[width=\textwidth]{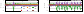}
    \caption{Handwash execution wrench: \gls{davla} makes a single clean press, whereas X-VLA\textsubscript{AFM} repeatedly presses and retracts (multipress) over a longer episode.}
    \label{fig:joint-force}
\end{figure*}

\subsubsection*{RQ4: Gated cross-attention preserves pretrained representations under new modalities.}
\label{sec:rq4}
X-VLA$_{AFM}$ has the same force and memory inputs as \gls{davla} but concatenates them into one flat token sequence. Despite identical information, it degrades across all tasks (54.3\% vs 95.2\%). Pushing unseen new tokens through pretrained self-attention disrupts the backbone's visual-language features. The policy reaches targets but cannot finish interactions, under-pressing on button, repeatedly pressing and retracting on handwash (Fig.\ref{fig:joint-force}), and stalling on whiteboard. \gls{gca} avoids this by adding new tokens as zero-initialized residuals, leaving pretrained weights untouched. The gap shows multimodal gains depend not just on the information itself, but how it enters a pretrained backbone as well.

\section{Limitations}

\gls{davla} uses high-frequency force to build better representations, but it does not use force to correct actions within a chunk. This leads to reduced performance on very contact-heavy tasks (e.g.\ $80\%$ on socket), where small alignment errors are not fixed mid-chunk. Using force to refine actions is the direct next step to improve these cases. The vision side is also only partly decoupled as the camera updates on a fixed timer rather than when the scene changes. A detector that triggers the \gls{vlm} on change would close this gap.

Finally our force signal comes from Franka's built-in joint-torque estimate, not a dedicated sensor. This shows the gains hold even without special hardware. Adding an external \gls{ft} sensor, torque-level control, and more modalities could push performance further.

\label{sec:limitation}

\section{Conclusion}
\label{sec:conclusion}

We introduced \gls{davla}, a \gls{vla} built on a simple principle: each modality should update and be remembered at its own natural sensor rate, not forced onto one shared clock. \gls{davla} keeps a separate latent buffer per modality, refreshes each at its sensor rate, and lets the action head read all of them continuously. New high-frequency modalities are added through gated cross-attention, so the pretrained backbone stays intact. Across seven contact-rich real-world tasks, this more than doubles the success rate of the strongest synchronous baseline (95.2\% vs.\ 40.95\%) while running smoothly and reactively at 100\,Hz. These results show that letting each sensor update the model at its own rate, instead of forcing everything onto one shared rate, is a practical way to build better manipulation policies. Next, we plan to use force to correct actions while they happen, and to add other fast sensors in the same way.

\acknowledgments{The research presented in this paper was
funded by the Deutsche Forschungsgemeinschaft (DFG, German Research Foundation) – 448648559. The authors gratefully acknowledge the computing time provided on the high-performance computer HoreKa by the National High-Performance Computing Center at KIT and by Gauss Centre for Supercomputing e.V. (www.gauss-centre.eu) for GCS Supercomputer JUPITER at Jülich Supercomputing Centre (JSC).}

\clearpage

\bibliography{example}  

\clearpage
\appendix
\clearpage
\begin{center}
    {\Large\bfseries Appendix}
\end{center}

\section{Robot Platform and Sensor Suite}
\label{app:robot_platform}

All real-world experiments are conducted on a \textbf{Franka Emika Panda} 7-DoF robot arm equipped with a \textbf{Robotiq 2F-85} parallel-jaw gripper. Our setup follows the DROID-style robot platform, using one external third-person camera and one wrist-mounted camera for visual observations. The sensor suite comprises:

\begin{itemize}[leftmargin=*, itemsep=2pt]
    \item \textbf{Third-person RGB camera}: External right-view RGB camera following the DROID camera setup, mounted at a fixed side viewpoint with respect to the workspace. The RGB stream is recorded at 25\,Hz, resized to 256$\times$256, and temporally upsampled to 100\,Hz for synchronized training with proprioceptive and force/torque observations.
    
    \item \textbf{Wrist-mounted RGB camera}: Wrist RGB camera attached near the robot end-effector, providing an egocentric view of the manipulation scene. The RGB stream is recorded at 25\,Hz, resized to 256$\times$256, and temporally upsampled to 100\,Hz for synchronized training with proprioceptive and force/torque observations.

    \item \textbf{Force/torque}: Force-related observations are obtained from the Franka's internal estimates rather than from an external force/torque sensor. The recorded 14-D force/torque observation consists of 7 external joint-torque estimates, a 6-D external wrench estimate, and the gripper current, all logged at 100\,Hz. No dedicated contact sensor was used (as in \cite{lodige2025use}). For our method, we use only 7-D external joint-torque estimates. 

    \item \textbf{Proprioception}: Proprioceptive observations consist of the 7 robot joint positions and the gripper state, forming an 8-D state vector recorded at 100\,Hz.
\end{itemize}

Each episode is stored in a LeRobot-style format with synchronized RGB observations, proprioceptive state, force/torque observations, actions, timestamps, and frame indices. The RGB cameras are recorded at 25\,Hz and temporally upsampled to 100\,Hz by holding the most recent visual observation, so that vision, proprioception, force/torque, and action streams are aligned within each training batch. The real-world dataset used in this work contains 50--60 episodes for each task.

The robot operates over a fixed tabletop workspace. All tasks use the same hardware configuration; no task-specific changes to sensor mounting are made between evaluations. For the 200\,Hz controller experiments, the same sensor suite is used with the \textit{libfranka} control loop running at 200\,Hz for proprioception and force/torque, while the camera streams remain at 25\,Hz.

\section{Training and Implementation Details}
\label{app:training_details}

\begin{table}[h]
\centering
\small
\begin{tabular}{ll}
\toprule
\textbf{Parameter} & \textbf{Value} \\
\midrule
Backbone & X-VLA \\
Learning rate & $2\times10^{-4}$ \\
Global Batch size & 192 \\
Training steps & 20{,}000 \\
Training hardware & NVIDIA GH200 480GB GPU node \\
Inference hardware & NVIDIA RTX 4060 Ti \\
Backbone training & Finetuned vision encoders, action experts\\
Visual input rate & 25\,Hz \\
Control rate & 100\,Hz \\
Force input rate & 100\,Hz \\
Proprioception input rate & 100\,Hz \\
Visual stride $S$ for history & 8 \\
GCA insertion points & Every 4th transformer layer of the action expert \\
\bottomrule
\end{tabular}
\caption{Training and implementation details for the reported experiments.}
\label{tab:training_details}
\end{table}

We train all policies using the same real-world demonstration data, sensor streams, and X-VLA backbone initialization. RGB observations are recorded at 25\,Hz and temporally aligned to the 100\,Hz control timeline used for proprioception, force observations, and actions. All policies are trained with two RGB views, 256$\times$256 image observations, 8-D proprioceptive state observations, and 7-D external joint-torque estimates from the Franka internal sensors. The policies are trained to predict 8-D actions (7-D joint positions and 1-D gripper state). During inference, visual tokens are computed once and cached. The VLM is queried to refresh them every 4 inference steps rather than at every control step.

\section{Failure Modes and Ablation Observations}
\label{app:failure_modes}

\paragraph{Synchronous baselines.}
X-VLA$_{25}$ generally exhibits jerky free-space motion. X-VLA$_{100}$ stalls mid-task on Scarf after reaching a visually stable intermediate configuration. The policy does not recover from this stall and fails to complete the remaining folding steps. On Sweep, X-VLA$_{100}$ loses the smooth continuous motion profile that X-VLA$_{25}$ produces at lower frequency. The redundant-frame bias causes the policy to issue small inconsistent commands instead of committing to a sustained sweeping motion.

\paragraph{\gls{davla} without force and memory.}
Separating visual encoding from the control loop avoids the redundant-frame problem. However, on whiteboard cleaning the policy can become slow or stall mid-trajectory without the temporal context needed to track progress across sequential steps.

\paragraph{Full \gls{davla}.}
Memory stabilizes long-horizon sequencing and prevents repeated interactions. Force provides the contact-state signal needed to terminate interactions at the right moment. This combination is especially important for handwash, Lego, and socket insertion, where success depends on both reaching the correct object and executing the final contact phase precisely.

\section{Motion Smoothness Metrics}
\label{app:smoothness_metrics}

We report two complementary metrics to evaluate trajectory quality on the 
Sweep task: spectral arc length (SPARC) for command smoothness and tracking lag for execution responsiveness. Each metric is suited to a different comparison, as described later. We focus on Sweep because it requires sustained, continuous arm motion across the full episode, making differences in command smoothness and tracking responsiveness most apparent. It is also the task where all configurations achieve measurable execution, enabling a complete cross-method comparison. The smoothness difference is most directly visible in the project videos. The metrics below provide quantitative support.

\paragraph{Spectral arc length (SPARC).}
SPARC measures the arc length of the normalized Fourier magnitude spectrum 
of the 7D joint-command trajectory. Smoother trajectories concentrate 
spectral energy at low frequencies and produce lower SPARC values. We 
evaluate all 100\,Hz configurations on their native command signals. 
\mbox{X-VLA$_{25}$} is excluded from this comparison: converting 25\,Hz 
commands to 100\,Hz via zero-order hold introduces staircase artifacts that 
inflate SPARC independently of motion quality, making cross-frequency 
comparison via SPARC meaningless.

As shown in Figure~\ref{fig:sparc_appendix}, among 100\,Hz methods, 
X-VLA$_{100}$ produces the highest SPARC value on Sweep, consistent with 
its redundant-frame bias causing small, inconsistent joint commands. 
\mbox{\gls{davla}} achieves the lowest SPARC, confirming it issues the 
smoothest commands among all 100\,Hz configurations.

\definecolor{davlacolor}{RGB}{0,114,178}
\definecolor{xvla100color}{RGB}{230,159,0}
\definecolor{xvla25color}{RGB}{0,158,115}
\definecolor{avgdemocolor}{RGB}{160,160,160}

\definecolor{xvlaafmcolor}{RGB}{86,166,75}
\definecolor{nofmcolor}{RGB}{230,85,85}
\definecolor{noforcecolor}{RGB}{176,124,160}
\definecolor{nomemorycolor}{RGB}{113,177,177}
\definecolor{ourscolor}{RGB}{230,190,80}

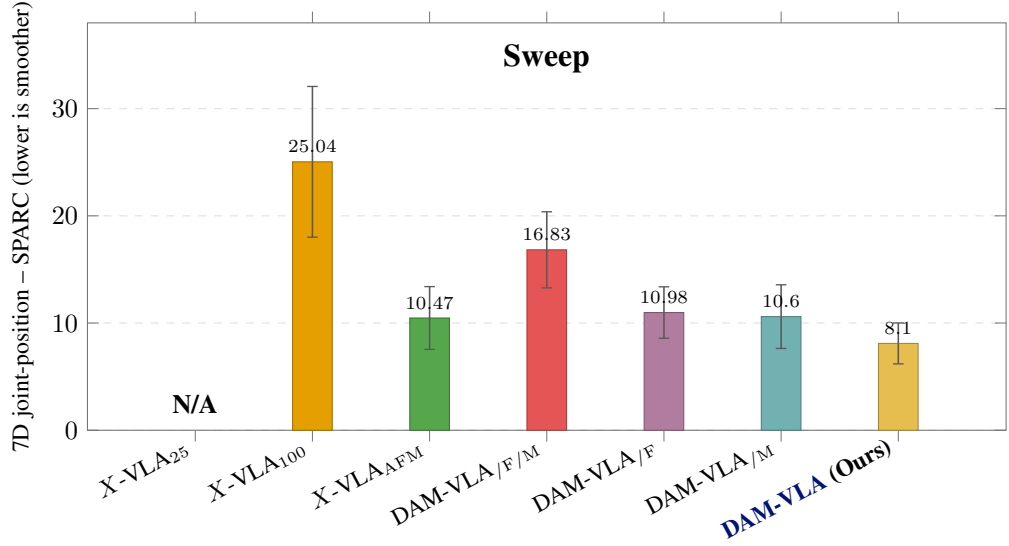
\begin{figure}[t]
    \centering
    \resizebox{0.95\columnwidth}{!}{%
    \begin{tikzpicture}
        \begin{axis}[
            width=13.8cm,
            height=7.0cm,
            ybar,
            bar width=15pt,
            ymin=0,
            ymax=38,
            xmin=0.5,
            xmax=7.5,
            enlarge x limits=0.06,
            xtick={1,2,3,4,5,6,7},
            xticklabels={
                {$X$-VLA$_{25}$},
                {$X$-VLA$_{100}$},
                {$X$-VLA$_{\mathrm{AFM}}$},
                {DAM-VLA$_{\mathrm{/F/M}}$},
                {DAM-VLA$_{\mathrm{/F}}$},
                {DAM-VLA$_{\mathrm{/M}}$},
                {\textbf{\gls{davla} (Ours)}}
            },
            x tick label style={rotate=28, anchor=east, font=\small},
            ylabel={7D joint-position -- SPARC (lower is smoother)},
            ylabel style={font=\small},
            ymajorgrids=true,
            grid style={gray!25, dashed},
            axis line style={black!55},
            tick style={black!55},
            title={SPARC 7D Vector Comparison},
            title style={font=\large\bfseries, yshift=5mm},
            clip=true,
            error bars/y dir=both,
            error bars/y explicit,
            error bars/error bar style={
                draw=black!65,
                line width=0.6pt
            },
            error bars/error mark=-,
            error bars/error mark options={
                rotate=90,
                draw=black!65,
                line width=0.5pt,
                mark size=2pt
            },
            nodes near coords,
            every node near coord/.append style={
                font=\scriptsize,
                /pgf/number format/fixed,
                /pgf/number format/precision=2,
                black
            }
        ]

        \node[
            font=\large\bfseries,
            anchor=north
        ] at (axis description cs:0.5,0.97) {Sweep};

        \node[font=\bfseries] at (axis cs:1,2.4) {N/A};

        \addplot+[
            fill=xvla100color,
            draw=xvla100color!70!black,
            bar shift=0pt
        ] coordinates {
            (2,25.04) +- (0,7.03)
        };

        \addplot+[
            fill=xvlaafmcolor,
            draw=xvlaafmcolor!70!black,
            bar shift=0pt
        ] coordinates {
            (3,10.47) +- (0,2.93)
        };

        \addplot+[
            fill=nofmcolor,
            draw=nofmcolor!70!black,
            bar shift=0pt
        ] coordinates {
            (4,16.83) +- (0,3.55)
        };

        \addplot+[
            fill=noforcecolor,
            draw=noforcecolor!70!black,
            bar shift=0pt
        ] coordinates {
            (5,10.98) +- (0,2.40)
        };

        \addplot+[
            fill=nomemorycolor,
            draw=nomemorycolor!70!black,
            bar shift=0pt
        ] coordinates {
            (6,10.60) +- (0,2.97)
        };

        \addplot+[
            fill=ourscolor,
            draw=ourscolor!70!black,
            bar shift=0pt
        ] coordinates {
            (7,8.10) +- (0,1.91)
        };

        \end{axis}
    \end{tikzpicture}%
    }
    \caption{
    Command smoothness on Sweep using 7D joint commands. Only
    100\,Hz configurations are shown. \mbox{$X$-VLA$_{25}$} is excluded
    because zero-order hold upsampling to 100\,Hz introduces staircase
    artifacts that inflate SPARC independently of motion quality.
    Lower values indicate smoother commands with less high-frequency content.
    }
    \label{fig:sparc_appendix}
\end{figure}

\paragraph{Tracking lag.}
Tracking lag measures the temporal delay between commanded and measured joint motion. For each joint, we compute command and measured velocities via finite differences. We then find the delay $\tau \in [0,\,0.5]$\,s at which the measured velocity best matches the command velocity, estimated via normalized cross-correlation. The lower bound of zero enforces causality. The robot can only lag behind the command, not ahead of it. The upper bound of 0.5\,s excludes delays beyond any plausible tracking range for the tasks studied, all observed values fall well below this ceiling. We report the mean lag across joints and episodes. Unlike SPARC, this metric requires no frequency normalization. All command signals are compared against the same measured joint signal, making it valid across all configurations including \mbox{X-VLA$_{25}$}.
While the per-step lag differences are modest, a persistent lag across 
multiple control steps compounds into visible trajectory deviation 
over a full episode.

\definecolor{davlacolor}{RGB}{0,114,178}
\definecolor{xvla100color}{RGB}{230,159,0}
\definecolor{xvla25color}{RGB}{0,158,115}
\definecolor{avgdemocolor}{RGB}{160,160,160}

\definecolor{xvlaafmcolor}{RGB}{86,166,75}
\definecolor{nofmcolor}{RGB}{230,85,85}
\definecolor{noforcecolor}{RGB}{176,124,160}
\definecolor{nomemorycolor}{RGB}{113,177,177}
\definecolor{ourscolor}{RGB}{230,190,80}

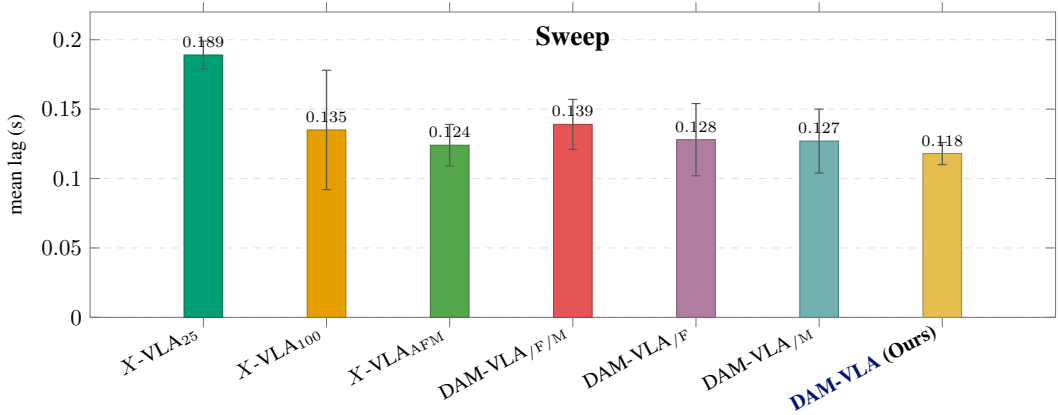
\begin{figure}[t]
    \centering
    \resizebox{\textwidth}{!}{%
    \begin{tikzpicture}
        \begin{axis}[
            width=16.5cm,
            height=6.3cm,
            ybar,
            bar width=17pt,
            ymin=0,
            ymax=0.22,
            xmin=0.5,
            xmax=7.5,
            enlarge x limits=0.06,
            xtick={1,2,3,4,5,6,7},
            xticklabels={
                {$X$-VLA$_{25}$},
                {$X$-VLA$_{100}$},
                {$X$-VLA$_{\mathrm{AFM}}$},
                {DAM-VLA$_{\mathrm{/F/M}}$},
                {DAM-VLA$_{\mathrm{/F}}$},
                {DAM-VLA$_{\mathrm{/M}}$},
                {\textbf{\gls{davla} (Ours)}}
            },
            x tick label style={rotate=28, anchor=east, font=\small},
            ylabel={mean lag (s)},
            ylabel style={font=\small},
            ytick={0,0.05,0.10,0.15,0.20},
            yticklabel style={
    /pgf/number format/fixed,
    /pgf/number format/precision=2
},
scaled y ticks=false,
            ymajorgrids=true,
            grid style={gray!25, dashed},
            axis line style={black!55},
            tick style={black!55},
            title={Tracking Lag Mean},
            title style={font=\large\bfseries, yshift=5mm},
            clip=true,
            error bars/y dir=both,
            error bars/y explicit,
            error bars/error bar style={
                draw=black!65,
                line width=0.6pt
            },
            error bars/error mark=-,
            error bars/error mark options={
                rotate=90,
                draw=black!65,
                line width=0.5pt,
                mark size=2pt
            },
            nodes near coords,
            every node near coord/.append style={
                font=\scriptsize,
                /pgf/number format/fixed,
                /pgf/number format/precision=3,
                black
            }
        ]

        \node[
            font=\large\bfseries,
            anchor=north
        ] at (axis description cs:0.5,0.98) {Sweep};

        \addplot+[
            fill=xvla25color,
            draw=xvla25color!70!black,
            bar shift=0pt
        ] coordinates {
            (1,0.189) +- (0,0.010)
        };

        \addplot+[
            fill=xvla100color,
            draw=xvla100color!70!black,
            bar shift=0pt
        ] coordinates {
            (2,0.135) +- (0,0.043)
        };

        \addplot+[
            fill=xvlaafmcolor,
            draw=xvlaafmcolor!70!black,
            bar shift=0pt
        ] coordinates {
            (3,0.124) +- (0,0.015)
        };

        \addplot+[
            fill=nofmcolor,
            draw=nofmcolor!70!black,
            bar shift=0pt
        ] coordinates {
            (4,0.139) +- (0,0.018)
        };

        \addplot+[
            fill=noforcecolor,
            draw=noforcecolor!70!black,
            bar shift=0pt
        ] coordinates {
            (5,0.128) +- (0,0.026)
        };

        \addplot+[
            fill=nomemorycolor,
            draw=nomemorycolor!70!black,
            bar shift=0pt
        ] coordinates {
            (6,0.127) +- (0,0.023)
        };

        \addplot+[
            fill=ourscolor,
            draw=ourscolor!70!black,
            bar shift=0pt
        ] coordinates {
            (7,0.118) +- (0,0.008)
        };

        \end{axis}
    \end{tikzpicture}%
    }
    \caption{
    Mean tracking lag comparison on Sweep. Tracking lag measures the estimated
    temporal delay between the commanded joint motion and the measured joint motion.
    Unlike SPARC, this metric is not affected by frequency mismatch and provides a fair
    comparison across all methods. Lower values indicate that the measured robot motion
    follows the command more promptly.
    }
    \label{fig:tracking_lag_appendix}
\end{figure}

With this in mind, the results in Figure~\ref{fig:tracking_lag_appendix} are consistent with our broader findings. X-VLA$_{25}$ shows the highest lag (0.189\,s), which is partly structural: at 25\,Hz, commands update every 40\,ms, so elevated lag is mechanically expected regardless of command quality. Among 100\,Hz methods, where all configurations share the same update period, \gls{davla} achieves the lowest lag (0.116\,s). This is consistent with its smoother command profile as measured by SPARC. Finally, \gls{davla} completes Sweep episodes faster than all other configurations (22.5\,s on average), ruling out the possibility that lower lag simply reflects slower, easier-to-follow motion.
Neither SPARC nor tracking lag is a perfect measure of motion quality in isolation, but together they tell a consistent story: \gls{davla} issues smoother commands among 100\,Hz methods and the robot follows them more promptly across all methods, in line with the qualitative behavior visible in the project videos.

\section{Episode Length and Execution Time}
\label{app:episode_length}
\definecolor{davlacolor}{RGB}{0,114,178}
\definecolor{xvla100color}{RGB}{230,159,0}
\definecolor{xvla25color}{RGB}{0,158,115}
\definecolor{avgdemocolor}{RGB}{160,160,160}

\begin{figure*}[t]
    \centering
    \resizebox{\textwidth}{!}{%
    \begin{tikzpicture}
        \begin{axis}[
            width=16.5cm,
            height=5.4cm,
            ybar,
            bar width=7pt,
            ymin=0,
            ymax=100,
            xmin=0.5,
            xmax=7.5,
            xtick={1,2,3,4,5,6,7},
            xticklabels={Sweep,Handwash,Lego,Scarf,Whiteboard,Button,Socket},
            x tick label style={rotate=18, anchor=east, font=\small},
            ylabel={Execution Time (s)},
            ylabel style={font=\small},
            ytick={0,20,40,60,80},
            ymajorgrids=true,
            grid style={gray!25, dashed},
            axis line style={black!55},
            tick style={black!55},
            title style={font=\large},
            legend style={
                at={(0.02,0.98)},
                anchor=north west,
                draw=gray!30,
                fill=white,
                font=\small
            },
            legend cell align={left},
            clip=false,
            error bars/y dir=both,
            error bars/y explicit,
            error bars/error bar style={
                draw=black!65,
                line width=0.6pt
            },
            error bars/error mark=-,
            error bars/error mark options={
                rotate=90,
                draw=black!65,
                line width=0.5pt,
                mark size=2pt
            }
        ]

        \addplot+[
            area legend,
            fill=davlacolor,
            draw=davlacolor!70!black,
            bar shift=-13.5pt
        ] coordinates {
            (1,23.35) +- (0,2.08)
            (2,13.33) +- (0,2.33)
            (3,28.87) +- (0,4.99)
            (4,25.13) +- (0,1.84)
            (5,24.38) +- (0,3.17)
            (6,10.77) +- (0,2.45)
            (7,22.30) +- (0,4.68)
        };
        \addlegendentry{\gls{davla} (Ours)}

        \addplot+[
            area legend,
            fill=xvla25color,
            draw=xvla25color!70!black,
            bar shift=-4.5pt
        ] coordinates {
            (1,25) +- (0,2)
            (4,28) +- (0,2)
            (5,25.20) +- (0,2)
            (6,16) +- (0,2)
            (7,80) +- (0,5)
        };
        \addlegendentry{$X$-VLA$_{25}$}

        \addplot+[
            fill=xvla25color!85,
            draw=xvla25color!70!black,
            bar shift=-4.5pt,
            postaction={pattern=north east lines, pattern color=black!70},
            forget plot
        ] coordinates {
            (2,15) +- (0,2)
            (3,44) +- (0,3)
        };

        \addplot+[
            area legend,
            fill=xvla100color,
            draw=xvla100color!70!black,
            bar shift=4.5pt
        ] coordinates {
            (1,50) +- (0,2)
            (4,37) +- (0,2)
            (5,83) +- (0,3)
            (6,20) +- (0,2)
        };
        \addlegendentry{$X$-VLA$_{100}$}

        \addplot+[
            fill=xvla100color!85,
            draw=xvla100color!70!black,
            bar shift=4.5pt,
            postaction={pattern=north east lines, pattern color=black!70},
            forget plot
        ] coordinates {
            (2,17) +- (0,2)
        };

        \addplot+[
            area legend,
            fill=avgdemocolor,
            draw=avgdemocolor!70!black,
            bar shift=13.5pt,
            error bars/y dir=none
        ] coordinates {
            (1,13.88)
            (2,7.37)
            (3,20.57)
            (4,13.60)
            (5,21.9)
            (6,6.45)
            (7,17.54)
        };
        \addlegendentry{Avg. demo time}

        \end{axis}
    \end{tikzpicture}%
    }
    \caption{Execution times across different tasks and model configurations. Hatched bars indicate partial or no proper success. Gray bars show the average demonstration time in the dataset. }
    \label{fig:task_execution_time}
\end{figure*}
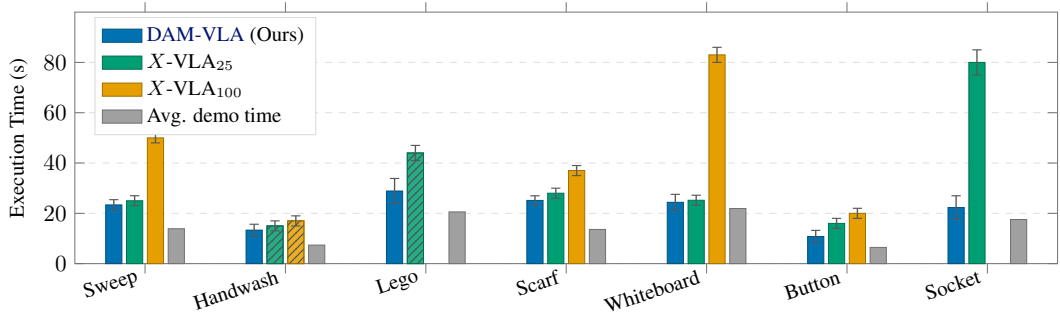
Beyond task success rate, episode duration offers a complementary view of policy quality. A policy that hesitates, stalls, or repeatedly retries will accumulate long episode lengths even on tasks it eventually completes. Conversely, short durations on successful tasks suggest the policy commits to actions confidently and without unnecessary repetition.
Figure~\ref{fig:task_execution_time} shows mean episode length across all seven tasks for \gls{davla}, X-VLA$_{25}$, and X-VLA$_{100}$, alongside the average human demonstration time from the dataset (gray bars) as a reference anchor. Hatched bars mark cases with zero or partial success. Episode lengths for these cases do \emph{not} reflect successful completion and should be interpreted accordingly. We include them as a best-effort record of how long each policy remained active before episode termination, so that comparisons across configurations remain as informative as possible.
\paragraph{Successful tasks.}
On tasks where all three configurations achieve measurable success, \gls{davla} consistently produces the shortest or near-shortest episode lengths. On Scarf, execution times are broadly comparable across configurations. On Sweep, \gls{davla} and X-VLA$_{25}$ finish at similar durations while X-VLA$_{100}$ takes noticeably longer, consistent with its tendency to issue overly conservative commands. The strongest difference appears on Whiteboard, where X-VLA$_{100}$ averages 83s, more than three times the 24.4s recorded for \gls{davla}, consistent with the stalling behavior described in Section~\ref{app:failure_modes}. On Button, \gls{davla} completes cleanly in 10.8s, well below the 16s and 20s recorded for X-VLA$_{25}$ and X-VLA$_{100}$ respectively, reflecting single-attempt execution without repeated contact probing.

\paragraph{Partial and failed cases.}
For Lego and Handwash, neither X-VLA$_{25}$ nor X-VLA$_{100}$ achieves any successful completion. The hatched bars represent the duration of active arm motion before episode termination. The policy made adjustments and attempted to reach the target but never completed the task. These times are reported for completeness and should not be interpreted as successful execution times. On Socket, X-VLA$_{25}$ records 80s corresponding to one near-successful attempt where the plug reached the socket but was slightly misaligned, preventing full insertion, with the remaining episodes timing out.
\paragraph{Overall.}
We note that the mean episode length for X-VLA$_{25}$ is heavily skewed by Socket (80s) and Lego (44s), both failed or partial cases. Excluding these, the average episode length for X-VLA$_{25}$ becomes comparable to \gls{davla}. This confirms that the advantage of \gls{davla} lies not in raw speed on individual tasks, but in maintaining reliable and consistent execution across the full task set, including tasks where the baselines fail entirely. Taken together with the smoothness metrics from Section~\ref{app:smoothness_metrics}, \gls{davla}'s shorter episode lengths reflect genuine decisiveness rather than faster or more aggressive motion.

\section{Replanning Frequency Ablation}
\label{app:replanning_ablation}
The main results reported in this paper use a 100\,Hz controller. To stress-test the limits of the asynchronous architecture, we additionally ran ablation experiments on the Handwash and Whiteboard tasks at 200\,Hz, varying the execution horizon $s$ to probe the trade-off between replanning frequency and motion quality.

In \gls{davla}, the input streams remain decoupled: force/torque and proprioception enter the buffer at high frequency, while visual tokens are updated sparsely and cached between updates. This means increasing the control frequency does not require more frequent visual inference, but it does affect how often the policy generates new action chunks. We compare two execution horizons (action chunk length): $s=22$, corresponding to an inference frequency of 8\,Hz, which is the default configuration used throughout the paper, and $s=6$, corresponding to 17\,Hz. Shorter horizons increase the replanning rate but reduce the amount of open-loop execution per generated chunk.

As shown in Table~\ref{tab:replanning_ablation}, both settings achieve full success on Handwash and Whiteboard. However, the two configurations differ in motion character. At $s=22$, the policy executes longer chunks and produces smooth, fluid motion. At $s=6$, the policy remains reactive but motion becomes visibly less fluid, as shorter chunks introduce more frequent transitions between generated sequences. Going beyond 17\,Hz, i.e.\ horizons shorter than $s=6$, caused task success to degrade, as even with sparse visual updates, the remaining VLM encoding latency becomes a bottleneck at very high replanning rates, leading to unstable execution.
\begin{table}[h]
\centering
\small
\begin{tabular}{lcclc}
\toprule
\textbf{Controller} & \textbf{Horizon} & \textbf{Freq.} & \textbf{Motion quality} & \textbf{Success} \\
\midrule
200\,Hz & $s=22$ & 8\,Hz & Smooth, fluid (default) & 100\% \\
200\,Hz & $s=6$ & 17\,Hz & Reactive, less fluid, slow task execution & 100\% \\
\bottomrule
\end{tabular}
\caption{Replanning frequency ablation for \gls{davla} under a 200\,Hz controller on Handwash and Whiteboard tasks. Execution horizon $s$ controls steps executed per generated chunk; shorter horizons increase replanning frequency.}
\label{tab:replanning_ablation}
\end{table}

\end{document}